\documentclass[conference]{IEEEtran}
\IEEEoverridecommandlockouts
\usepackage{cite}
\usepackage{amsmath,amssymb,amsfonts}
\usepackage{algorithm}
\usepackage{algpseudocode}
\usepackage{graphicx}
\usepackage{textcomp}
\usepackage{xcolor}
\usepackage{algorithm}
\usepackage{algpseudocode}
\usepackage{booktabs}
\usepackage{adjustbox}

\def\BibTeX{{\rm B\kern-.05em{\sc i\kern-.025em b}\kern-.08em
    T\kern-.1667em\lower.7ex\hbox{E}\kern-.125emX}}
\begin{document}

\title{Hybrid Approach for Driver Behavior Analysis with Machine Learning, Feature Optimization, and Explainable AI\\
}


\author{
\IEEEauthorblockN{Mehedi Hasan Shuvo}
\IEEEauthorblockA{\textit{Department of CSE, DUET} \\
Gazipur, Bangladesh \\
mehedihasanshuvo.mail@gmail.com}
\and
\IEEEauthorblockN{Md. Raihan Tapader}
\IEEEauthorblockA{\textit{Department of CSE, DUET} \\
Gazipur, Bangladesh \\
204006@student.duet.ac.bd}
\and
\IEEEauthorblockN{Nur Mohammad Tamjid}
\IEEEauthorblockA{\textit{Department of CSE, DUET} \\
Gazipur, Bangladesh \\
tamjid.cse.duet@gmail.com}
\and
\IEEEauthorblockN{Sajjadul Islam}
\IEEEauthorblockA{\textit{Department of CSE, DUET} \\
Gazipur, Bangladesh \\
2204111@student.duet.ac.bd}
\and
\IEEEauthorblockN{Ahnaf Atef Choudhury}
\IEEEauthorblockA{\textit{Department of IST, George Mason University} \\
Fairfax, USA \\
achoudh9@gmu.edu}
\and
\IEEEauthorblockN{Jia Uddin}
\IEEEauthorblockA{\textit{AI \& Big Data Department, Woosong University} \\
Daejeon, South Korea \\
jia.uddin@wsu.ac.kr}
}

\maketitle

\begin{abstract}
Progressive driver behavior analytics is crucial for improving road safety and mitigating the issues caused by aggressive or inattentive driving. Previous studies have employed machine learning and deep learning techniques, which often result in low feature optimization, thereby compromising both high performance and interpretability. To fill these voids, this paper proposes a hybrid approach to driver behavior analysis that uses a 12,857-row and 18-column data set taken from Kaggle. After applying preprocessing techniques such as label encoding, random oversampling, and standard scaling, 13 machine learning algorithms were tested. The Random Forest Classifier achieved an accuracy of 95\%. After deploying the LIME technique in XAI, the top 10 features with the most significant positive and negative influence on accuracy were identified, and the same algorithms were retrained. The accuracy of the Random Forest Classifier decreased slightly to 94.2\%, confirming that the efficiency of the model can be improved without sacrificing performance. This hybrid model can provide a return on investment in terms of the predictive power and explainability of the driver behavior process. 
\end{abstract}


\section{Introduction}

Driver behavior analysis has become a critical area of research due to its direct impact on road safety, traffic management, and accident prevention. Dangerous, careless, and erratic driving habits often cause more serious accidents that lead to both human and financial loss. Machine learning and XAI have already opened new avenues for understanding and predicting driver behavior. Yet, many current methods encounter feature optimization and model explainability challenges, frequently compromising prediction effectiveness and repeatability. Numerous studies have investigated machine learning techniques for analyzing and interpreting driver behavior to enhance road safety and optimize driving performance.

Amer et al. \cite{10418298} proposed a multi-label classification approach for aggressive driving detection using SVM and random forest classifiers; also, their finding was 9.94\% (SVM) and 8.68\% (RF) through feature selection. On the other hand, using transfer learning techniques, including MLP, CNN, LSTM, ResNet, and MobileNet, to predict injury severity in traffic accidents was proposed in \cite{10477984}. Their finding: MobileNet achieved the highest accuracy of 98.17\%.
Detecting aggressive driving on horizontal curves based on TLC and motion-related variables proposed in \cite{8169689,abou2020application,lindow2020ai}, they also used a random forest classifier. However, they did not explore unsupervised learning, monitoring optimization, adaptation to diverse driving environments, or real-time personalization, presenting opportunities for future advancements.

In \cite{ghandour2021driver}, the authors analyze driver behavior (normal, drowsy, aggressive) using machine learning models such as logistic regression, random forest, neural networks, and gradient boosting on real-world datasets. Gradient boosting emerged with the highest accuracy and precision, while neural networks underperformed. However, the study has not yet incorporated road speed limits and mental workload. For advanced detection systems in \cite{nazat2024xai,islam2023interpretable,chengula2024enhancing,kuznietsov2024explainable}, authors proposed an XAI framework for anomaly detection in autonomous driving systems (ADS), using SHAP and LIME methods for interpretability and novel feature selection techniques to enhance performance. Their methodology, AdaBoost, achieved 87\% accuracy in detecting anomalies.

Using a denoising stacked autoencoder trained on GPS-based experimental driving test data proposed in \cite{bichicchi2020analysis,jain2024comprehensive}, an advanced unsupervised deep learning model generates RGB output layers to investigate the complex interactions between road environments and driver behavior. On the other hand, road accident frequency and driving risk using a naturalistic driving dataset of 77,859 km explored the relationship between driving context combinations and risk \cite{masello2023using}. They utilized XGBoost and Random Forest; key features such as speed limits, weather conditions, wind speed, traffic conditions, and road slope are ranked through SAE, providing critical insights for road safety stakeholders and insurers. Controller Area Network (CAN) bus signals for classifying driver behavior were proposed in \cite{fugiglando2018driving}. They analyzed data from 54 participants across 2000 trips and achieved robust clustering with up to 99\% data reduction without performance loss. In addition, ML and deep learning platforms for classifying eco-friendly driving behavior from vehicular data streams note that while both approaches are practical, proposed in \cite{peppes2021driving}, DL requires more execution time.

Chan et al. \cite{valente2024using} proposed a “DriverAlert” mobile app to analyze driver behavior using phone sensors and machine learning. K-Means excels in classification, and RF and SVM perform best in predictions. However, they expand only the data collection section in the app and refine models for greater accuracy.
On the other hand, XAI approaches for autonomous vehicles identified gaps in transparency and trust \cite{atakishiyev2024explainable}, where the authors suggested directions to enhance real-time decision explainability and societal acceptance, ultimately advancing safe and trustworthy autonomous driving technologies.

Detecting unusual driving behaviors such as fatigue, distraction, and reckless driving using smartphones; examining sensing techniques, detection algorithms, and challenges like lighting variations and sensor noise while suggesting solutions like context-aware systems and cloud updates proposed in \cite{chan2019comprehensive}. In the same way, the driver identification method \cite{lin2018driver,hwang2018apply} is based on unique driving behaviors, including acceleration, steering, and lane changes, achieving 100\% accuracy in tests by creating "driving fingerprints." Human-Centric Intelligent Driver Assistance System \cite{mccall2007driver} integrates driver, vehicle, and environmental data to enhance braking assistance, predict driver intent, and reduce distractions. These approaches demonstrate significant potential for creating safer, more efficient, intelligent driving systems.

\begin{table*}[h!]
\caption{Comparative Review of Driver Behavior Analysis Techniques in Machine Learning, Feature Optimization, and Explainable AI}
\label{comparison}
    \centering
    \begin{tabular}{cp{8cm}p{8.5cm}}
    \hline\\
    
        \cite{bouhsissin2024enhancing} & They utilize 10 different feature-selection techniques (filter, wrapper, and embedded) on the UAH-DriveSet to identify key features. Their Random Forest model achieves approximately 96.4\% accuracy with backward feature selection.  & They do not use explainability techniques (e.g., LIME/SHAP) to interpret individual predictions, nor do they examine retraining with top features after explanation. Their approach is not “hybrid” in combining feature optimization and XAI.\\
         
         \cite{khalid2025impact} & Comparative study using feature reduction methods (mutual information, PCA, SVD) across classifiers for driving behavior prediction; examines tradeoffs between dimensionality and accuracy. & They lack explainable AI for interpretability and do not explore how feature choice impacts transparency. They also don't propose a hybrid pipeline that incorporates explanation feedback into feature re-selection.\\
         
        \cite{qu2024comprehensive} & A review of vision-based driver behavior methods (facial expressions, posture, hand position) with ML, examining datasets, methods, and pros and cons across modalities. & The survey omits detailed discussion on tabular sensor methods, combining feature optimization with XAI in non-vision data, and lacks empirical comparisons of hybrid pipelines.\\
        
        \cite{mobini2025state} & Provides an overview of clustering, classification, hybrid modeling, and challenges in driver behavior analysis (e.g., data heterogeneity, imbalance, real-time constraints). & Since it is a review, it doesn't present new experimental results or test explainability. It highlights the need for interpretable models but doesn't propose specific hybrid pipelines.\\
         
         \cite{shabani2025innovative} & Proposes a system that combines Personalized Federated Learning (PFL), Digital Twin, Crowdsensing, and XAI for analyzing driving behavior, focusing on privacy and interpretability. & This is primarily a conceptual and architectural framework; it lacks empirical validation on datasets and does not thoroughly address retraining reduced-feature models after explanation.\\

         \cite{chengula2024enhancing} & Uses explainable AI to interpret ensemble models in driver behavior (in-vehicle camera-based) to reveal decision rationale in ADAS. & Focus on vision/ensemble models for camera input, not on tabular sensor data; it also doesn't link explanations to feature re-optimization.\\

        \cite{uddin2025abnormal} & Proposes a hybrid approach combining ML (tabular data) and DL (images/pose) to detect erratic driving behaviors by integrating different modalities. & They lack an explainability module (LIME/SHAP) for the tabular part or a feedback loop for retraining with top features. Their hybrid model combines image and tabular data, while yours is solely tabular with explanation feedback.\\
         
         \hline
    \end{tabular}
\end{table*}

AI-driven driver behavior analysis to optimize fuel consumption by analyzing patterns like acceleration and braking, emphasizing challenges such as data privacy and reliability in \cite{haghshenas2024artificial} while highlighting opportunities for eco-friendly driving and personalized training, focusing on advancing AI for sustainable transportation, minimizing fuel consumption, and reducing environmental impact. On the other hand, automated driving styles proposed in \cite{peintner2024driving}, which are aggressive and defensive, affect user trust and acceptance in autonomous vehicles, with a simulation study involving 49 participants revealing that driving style preferences vary by traffic scenario. However, their systems dynamically adjust driving styles using machine learning to enhance user satisfaction and trust in AVs. The authors \cite{zou2022vehicle} introduced an MHMM-based model to segment driving behavior and group similar drivers for vehicle acceleration prediction. GRU demonstrated superior performance over LSTM, reducing MAE by 38.8\% and offering better efficiency, particularly for homogeneous driver groups. Future research will validate this method by applying it to traffic data from other regions. A review of recent literature was conducted to understand current trends and limitations in machine learning–based driver behavior analysis. Table \ref{comparison} summarizes the main contributions and drawbacks of relevant studies.

This study makes several contributions. Our analysis of driver behavior strikes a balance between accuracy and interoperability. 

We make the following key contributions:
\begin{itemize}
\item We created a clean preprocessing pipeline, which included label encoding, followed by random oversampling to balance the dataset, sorted by the class label column, and using for each attribute the standard scale for feature engineering.

\item We evaluated 13 different machine learning algorithms used in our study, with the Random Forest Classifier yielding the best result with 95\% accuracy.

\item We implemented interpretation of ML predictions based on the LIME technique, such that strong impact features can be recognized, which in turn leads to interpretability of the models.

\item Finally, all the trained models were retrained after removing deduced prediction features from the cloned training set to show that interpretability does not come with a heavy price.

\end{itemize}
 The integration of these contributions forms a solid basis for driver behavior analysis based on explainability merits, bringing practical contributions to the domains of road safety and traffic.

The rest of this paper is structured as follows: Section II states the methodology, providing the dataset, preprocessing methods, and machine learning models. Section III presents the results and discussions, analyzing model performance and insights provided by explainable AI techniques. Finally, Section IV concludes the paper by summarizing key findings and outlining future research directions.

\section{Methodology}

\begin{figure}
    \centering
    \includegraphics[width=1\linewidth]{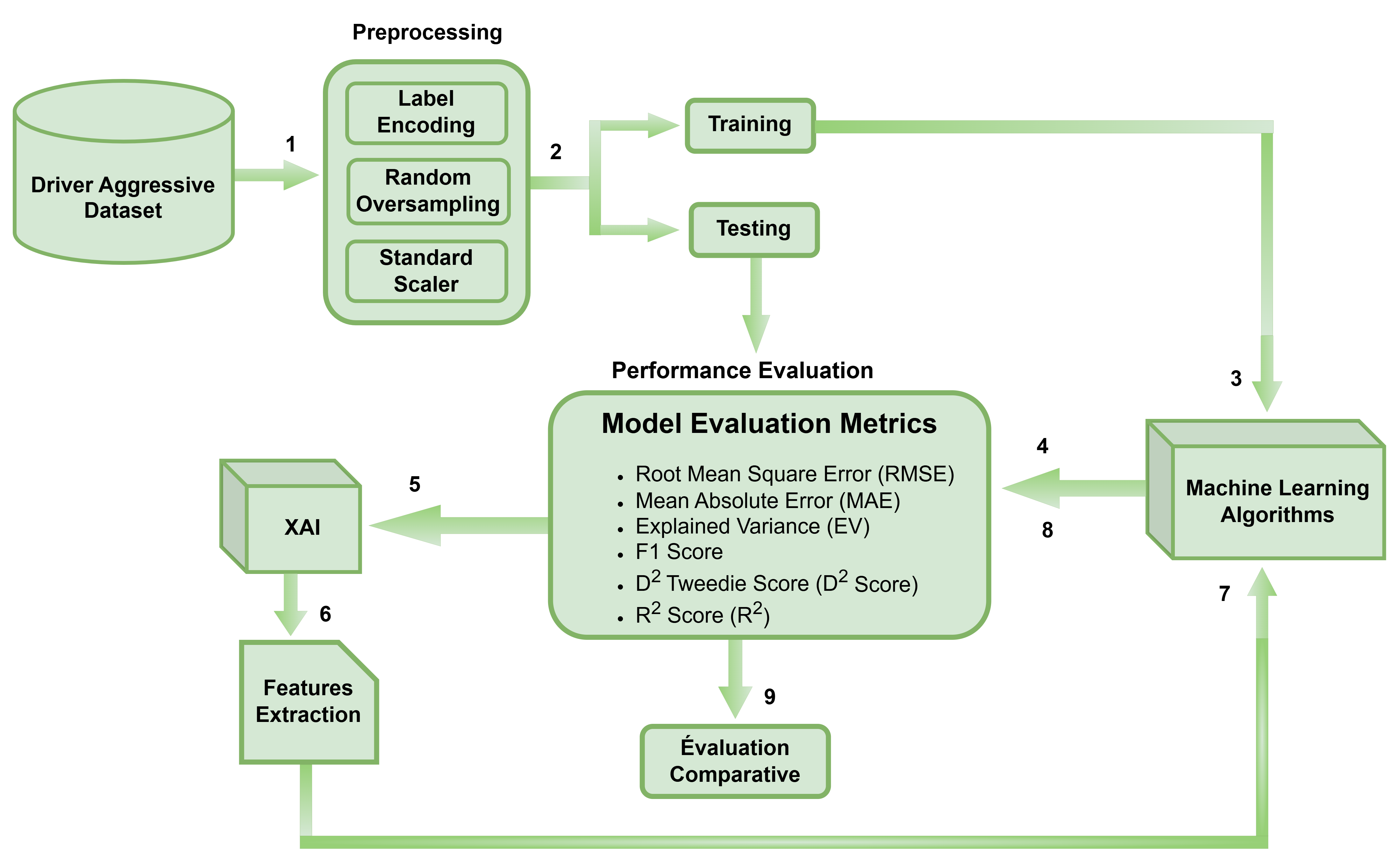}
    \caption{Proposed Methodology}
    \label{fig:proposed_methodology}
\end{figure}

In Figure \ref{fig:proposed_methodology}, we demonstrated our proposed methodology, dividing our methods into the following step-by-step procedure. Step 1: The Driver Aggressive Dataset is the input for the given processes. Step 2: The dataset will use the following preprocessing techniques: label encoding, intent oversampling, and standard scaling. After preprocessing, as mentioned in Step 3, the data is divided into training and testing sets for training machine learning models. Step 4: 13 machine learning algorithms are distributed to the training data, and their performance is assessed based on RMSE, MAE, EV, F1 Score,  $R^2$ Score, and $D^2$ Score.

Machine learning models perform weather predicting in each of the 5 steps, where Step 5 utilizes XAI techniques to know the importance of model features and model explainability. Step 6 is all about feature extraction, as the most impactful features that help predict model output are selected for further enhancement using tips on XAI. The extracted features can also be used in Step 7 to retrain the machine learning algorithms, enhancing their interpretability and performance. In Step 8, the retrained models are re-evaluated to ensure the optimized models work. Lastly, a comparative evaluation is carried out in Step 9 to determine if the model improvements made any difference and if the overall methodology appears to be working. This entire framework helps achieve a delicate balance between prediction accuracy and a baseline for model explainability. In Algorithm \ref{algo}, we also demonstrated our hybrid methodology for analyzing driver behavior.

\begin{algorithm}
\small
\caption{Hybrid Approach for Driver Behavior Analysis}
\label{algo}
\begin{algorithmic}[1] 
    \State \textbf{Step 1: Load and Preprocess Data}
    \State Load dataset from CSV file
    \State Handle missing values (fill with mean or drop)
    \For{each categorical column in dataset}
        \State Apply Label Encoding
    \EndFor
    \State Apply Random Oversampling to balance classes
    \State Normalize data using StandardScaler
    \State Split dataset into \texttt{(X\_train, X\_test, y\_train, y\_test)}
    
    \State 
    
    \State \textbf{Step 2: Train Machine Learning Models}
    \State Initialize \texttt{model\_list} with: Logistic Regression, Decision Tree, Random Forest, etc.
    \For{each model in \texttt{model\_list}}
        \State Train model using \texttt{X\_train, y\_train}
        \State Predict \texttt{y\_pred} using \texttt{X\_test}
        \State Compute Accuracy and F1-score
        \State Print model name, accuracy, and F1-score
    \EndFor
    
    \State 
    
    \State \textbf{Step 3: Select Best Model}
    \State Choose model with highest accuracy and F1-score

    \State 
    
    \State \textbf{Step 4: Explainability using XAI (LIME)}
    \State Initialize LIME Explainer
    \State Choose a test sample from \texttt{X\_test}
    \State Explain model prediction using LIME
    \State Identify top 10 important features

    \State 
    
    \State \textbf{Step 5: Feature Selection \& Retraining}
    \State Select Top 10 Features from feature importance
    \State Remove less important features from dataset
    \State Retrain best model using selected features
    \State Compute new accuracy and F1-score

    \State 
    
    \State \textbf{Step 6: Comparative Evaluation}
    \State \textbf{Display:} "Original Model - Accuracy:", original\_accuracy
    \State \textbf{Display:} "Optimized Model - Accuracy:", optimized\_accuracy
\end{algorithmic}
\end{algorithm}

\subsection{Dataset}
The Aggressive Driving Dataset was initially utilized, comprising three distinct components: \textit{Train.csv}, \textit{Train\_VehicleTravellingData.csv}, and \textit{Train\_WeatherData.csv}. These datasets contained $(12, 994, 5)$, $(162, 566, 10)$, and $(162,566, 9)$ entries and features, respectively, resulting in a total of 338,126 samples and 24 features. The data was sourced from Kaggle, and the feature names were customized to suit the specific requirements of this study. Through preprocessing and integration, the datasets were consolidated to produce a final dataset with 12,857 samples and 18 features. The final selected features were identified as critical factors influencing accident severity. These features include 'road conditions', 'driving style', 'vehicle length', 'vehicle speed', 'vehicle type', 'air temperature', 'Lighting condition', etc. This curated feature set provides a robust framework for exploring the relationship between aggressive driving behavior and accident severity.

\subsection{Preprocessing}

\begin{figure}
    \centering
    \includegraphics[width=1\linewidth]{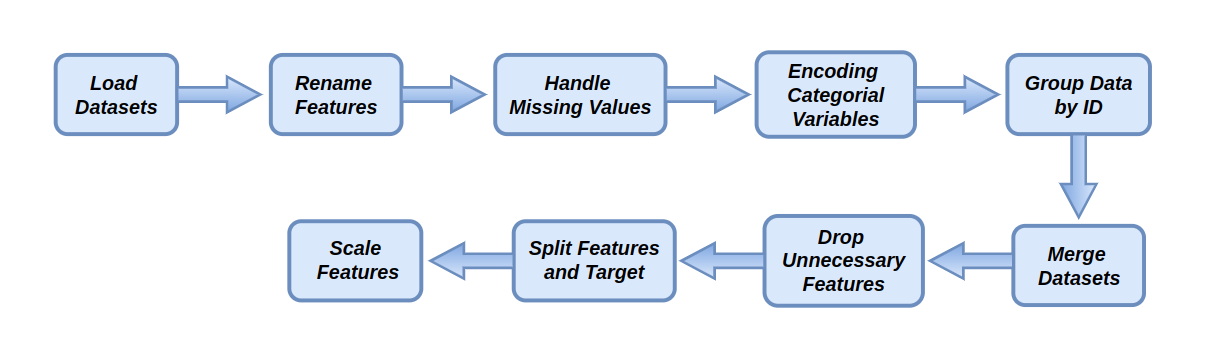}
    \caption{Data Processing}
    \label{fig:preprocessing}
\end{figure}

We have preprocessed our datasets in this research and used label encoding, random oversampling, and standard scaling techniques to prepare the final data, as demonstrated in Figure \ref{fig:preprocessing}. All these processes help transform the necessary data, including fixing class imbalance and normalizing the values of features to improve the model's prediction. First, the categorical data is converted into numerical value through label encoding. This allows for categorical variables just like the model can work. Then, random oversampling is used to resolve class imbalance by using synthetic copies of minority classes to balance class distribution on the classes. Finally, the standard scaler is applied to scale down the data and ensure all features have a mean of std 0 to 1. For improving performance by using normalized features, all the features contribute equally to the model. Combining the preprocessing techniques results in data properly formatted for model training, ultimately leading to more accurate and reliable results.

\subsection{ML Algorithms}

Machine learning algorithms form the basis of predictive models as they learn patterns from data to make predictions and decisions based on the patterns learned. These algorithms fall broadly under supervised, unsupervised, and reinforcement learning. In this approach, we selected supervised learning algorithms in which the model learns from the labeled data set to predict the outcomes. We explored algorithms based on decision trees, ensemble methods like random forests and gradient boosting, and classical linear models (logistic regression). Every algorithm has strengths, like dealing with non-linear relationships, transforming high-dimensional data, or providing model interpretability. They have achieved great popularity due to their high accuracy and robustness. Random Forest and Gradient Boosting algorithms handle complex datasets well because they combine the strengths of several weaker learners \cite{yadav2024impact,dabral2024machine, begum2025mlrec}. In contrast, simpler models such as logistic regression and naïve Bayes are often favored for their simplicity and high speed, mainly when interpretability is crucial. Hence, choosing the algorithm based on the problem to solve and the data set is essential.

\subsection{Evaluation Metrics}
\textbf{Accuracy:} Accuracy is a widely used evaluation metric to compare the performance of different classification models. It measures the frequency of correct positive and negative predictions for all instances of the event. To assess the effectiveness of various approaches, three evaluation metrics were employed, expressed using the terms total number of observations (T), False Positives (FP), True Negatives (TN), and True Positives (TP). Accuracy is defined as

\[
\text{Accuracy} = \frac{TP + TN}{T}
\]

\textbf{Precision:} This statistic is used to determine which forecasts were accurate in favor of the event. It provides a metric to evaluate a person's ability to foresee events accurately \cite{hossain_hybrid_2025}.

\[
 \text{Precision} = \frac{TP}{TP + FP } 
\]

\textbf{Recall:} This statistic is used to determine which forecasts were successful out of all actual occurrences of the event. It provides a way to evaluate how accurate predictions are for an event concerning each occurrence.

\[
 \text{Recall}= \frac{TP}{TP + FN } 
\]

\textbf{F1-score:} The F1-score is calculated by averaging the weighted precision and recall data. It is the best metric to collectively average and balance all evaluation metrics, as defined below.

\[
  \text{F1} = 2 \times \frac{\text{Precision} \times \text{Recall}}{\text{Precision} + \text{Recall}}
\]

\textbf{MSE:} Mean Squared Error represents the average squared difference between the original and predicted values in the data set. It measures the variance of the residuals. A perfect prediction is represented by an MSE of 0, and smaller values indicate increased model accuracy \cite{10499820}. A calculation of MSE is

\[
\text{MSE} = \frac{1}{N} \sum_{i=1}^{N} (y_{\text{act}} - y_{\text{pre}})^2
\]

\textbf{RMSE:} The RMSE measures \cite{jierula2021study} the vertical distance between the expected and actual values, quantifying the average error between the two. It is the MSE square root, and its range is \( (0, +\infty) \). A smaller RMSE indicates greater accuracy of the model. RMSE is more straightforward to grasp because it keeps the original units, unlike MSE.

\[
\text{RMSE} = \sqrt{\text{MSE}} = \sqrt{\frac{1}{N} \sum_{i=1}^{N}(y_{\text{act}} - y_{\text{pre}})^2}
\]

\textbf{R² Score:} The coefficient of determination, or R-squared, represents the proportion of variance in the dependent variable that is accounted for by the linear regression model. As a scale-free metric, R-squared is always less than or equal to one, ensuring its applicability regardless of the magnitude of the observed values.
\[
R^2 = \frac{SSR}{SST} = \frac{\sum_{i=1}^{N}(y_{\text{ori}} - y_{\text{pre}})^2}{\sum_{i=1}^{N}(y_{\text{ori}} - y_{\text{mean}})^2}
\]
  
\textbf{Explained Variance (EV):} Explained Variance (EV) is a metric used in statistical analysis to assess how much of the variability in the dependent variable can be explained by the independent variables in a regression model. The EV is expressed as a value between 0 and 1. The EV is typically derived from the \( R^2 \) statistic \cite{begum2023software}.

\[
\text{EV}(y_i, \hat{y}_i) = 1 - \frac{Var(y_i - \hat{y}_i)}{Var(y_i)}
\]

Here, it \( Var(y_i - \hat{y}_i) \) represents the difference between the actual and predicted values, and \( Var(y_i) \) represents the variance of the actual values. An EV near 1 signifies a strong model fit, while a value close to 0 indicates poor explanatory power.

\textbf{D² Score:} The \( D^2 \) score is used to evaluate how well models fit various probability distributions, including scaled Poisson, inverse Gaussian, gamma, and routine. It also includes Tweedie distributions, commonly used in generalized linear models (GLMs), and compound Poisson-gamma distributions, which have a positive probability mass at zero.

We present the average outcomes for twenty stratified shuffle splits for each experiment, using 12\% of the data for testing and 88\% for training.

\subsection{Expandable Artificial Intelligence (XAI)}
To ensure transparency in driver behavior analysis, we employ Explainable Artificial Intelligence (XAI) \cite{wormald2025abstracting} techniques to interpret the decision-making process of the models and provide insights into the factors influencing their predictions. Applying the LIME framework, we analyzed the predictions of our model to detect and classify different driving behaviors, such as aggressive, normal, and vague driving.  This interpretability is crucial for fostering trust in the model, especially in safety-critical applications such as driver behavior analysis, where understanding the reasoning behind predictions is essential to ensure accountability and informed decision-making.

\subsubsection{LIME}
The Explainable AI (XAI) technique known as LIME (Local Interpretable Model-Agnostic Explanations) \cite{mehta2022social} is intended to offer instance-specific, interpretable justifications for predictions generated by black-box models. Local fidelity is attained by utilizing surrogate models, such as Lasso or decision trees, to approximate the model's behavior close to a particular data point. LIME determines the most significant features influencing the prediction by applying feature selection algorithms and allocating weights to the data points. It is model-independent and can handle a variety of data formats, such as text, pictures, and videos. LIME is beneficial in fields like driver behavior analysis or applications where layman interpretability is crucial because of its clear, understandable explanations. The LIME framework allowed us to identify the most critical features contributing to the model's decisions, offering a clear understanding of how specific input data influenced the classification outcomes.



\section{Results and Discussion}

The experimental finding underscores the efficacy of the hybrid approach for driver behavior analysis. Performance metrics, including accuracy, F1 score, EV, MSE, RMSE, $R^2$, and $D^2$ scores, were evaluated using XAI techniques for 13 ML models before and after feature optimization. LGC and ABC reached the top performance with an accuracy of 69.7\% and an F1 score of 67.1\% before feature optimization. The RFC, in comparison, showed promising balanced performance at 65.7\% accuracy. Still, models such as KNN and MNB performed substantially less effectively, with MNB delivering the worst accuracy of 37.9\% demonstrated in Table \ref{tab:before_feature_extraction}.

\begin{figure}
    \centering
    \includegraphics[width=1\linewidth]{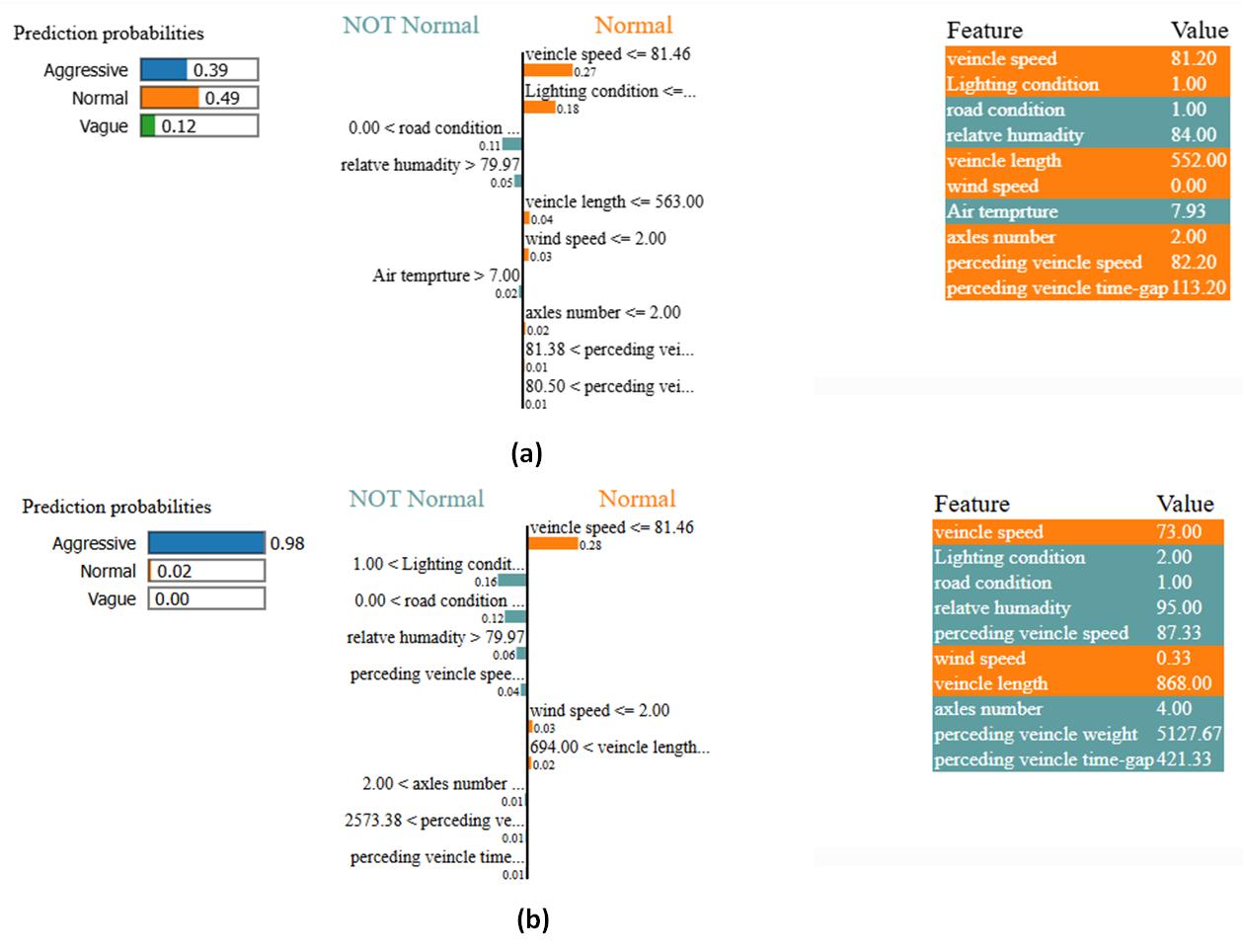}
    \caption{Features Importance}
    \label{fig:lime_results}
\end{figure}

\begin{figure}
    \centering
    \includegraphics[width=1\linewidth]{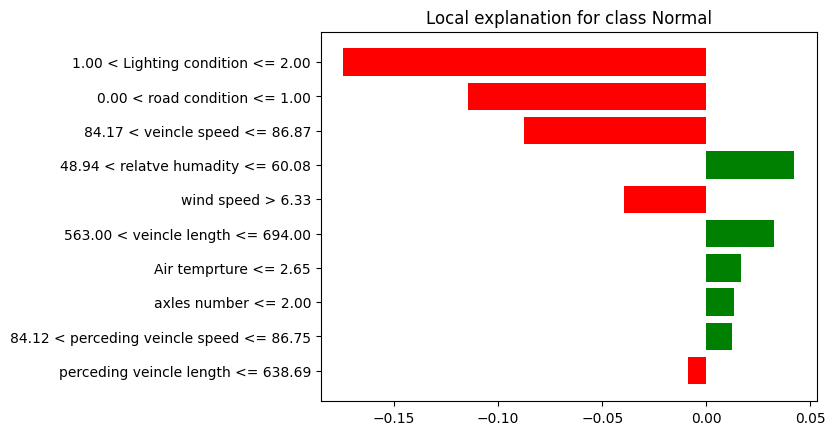}
    \caption{Local explanation for class Normal}
    \label{fig:lime_normal}
\end{figure}

Feature visualizations, shown in Figure \ref{fig:lime_results}, demonstrated the importance of features and local explanations for individual classes within specific data, as shown in Figure \ref{fig:lime_normal}, which added to the interpretability of the model and provided insight into how the model made predictions. Univariate feature selection reduced the original dataset of 12,857 rows and 18 columns to several features best describing the dataset without unnecessary information. This method helped lower the data's dimensionality and increase model interpretability. The LIME analysis revealed that environmental factors such as relative humidity, road conditions, and lighting conditions significantly affected drivers behavior based on the RFC feature importance scores. Also, vehicle specifications, such as 'vehicle speed', 'vehicle length', and 'preceding vehicle speed', were indispensable for learning interactions among the vehicles on the road. Incorporating these features into the model allowed the system to achieve high accuracy and produced valuable information about the driving environment and behavior.

\begin{table}[h!]
\centering
\caption{Model Performance Metrics Before Features Extraction}
\label{tab:before_feature_extraction}
\begin{tabular}{@{}p{1cm}p{0.7cm}p{0.7cm}ccccp{1cm}@{}}
\toprule
Model Name & Accuracy & F1 Score & EV & MSE & RMSE & $R^2$ & $D^2$ Score \\ \midrule
LR  & 0.528 & 0.512 & -0.485 & 0.562 & 0.744 & -0.487 & -0.487 \\
DTC & 0.567 & 0.528 & -0.401 & 0.527 & 0.700 & -0.401 & -0.401 \\
ABC & 0.696 & \textbf{0.671} & -0.043 & 0.397 & 0.531 & -0.062 & -0.062 \\
GBC & 0.629 & 0.666 & -0.026 & 0.379 & 0.523 & -0.047 & -0.047 \\
ETC & 0.639 & 0.605 & -0.155 & 0.434 & 0.519 & -0.163 & -0.163 \\
RFC & 0.657 & 0.629 & -0.070 & 0.410 & 0.545 & -0.090 & -0.090 \\
KNN & 0.479 & 0.435 & -0.464 & 0.600 & 0.760 & -0.522 & -0.522 \\
NB  & 0.601 & 0.588 & -0.417 & 0.504 & 0.714 & -0.430 & -0.430 \\
LDA & 0.634 & 0.604 & -0.270 & 0.455 & 0.673 & -0.271 & -0.271 \\
QDA & 0.587 & 0.556 & -0.482 & 0.526 & 0.750 & -0.501 & -0.501 \\
LGC & \textbf{0.697} & \textbf{0.671} & -0.006 & 0.373 & 0.514 & -0.029 & -0.029 \\
CBC & 0.689 & 0.660 & -0.002 & 0.379 & 0.515 & -0.030 & -0.030 \\
MNB & 0.379 & 0.370 & -0.926 & 0.736 & 1.005 & -1.010 & -1.010 \\ \bottomrule
\end{tabular}
\end{table}

\begin{table}[h!]
\centering
\caption{Model Performance Metrics After Features Extraction}
\label{tab:after_feature_extraction}

\begin{tabular}{@{}p{1cm}p{0.7cm}p{0.7cm}ccccp{1cm}@{}}
\toprule
Model Name & Accuracy & F1 Score & EV & MSE & RMSE & $R^2$ & $D^2$ Score \\ \midrule
LR  & 0.506 & 0.554 & -0.354 & 0.562 & 0.903 & -0.356 & -0.356 \\
DTC & 0.817 & 0.812 & 0.624  & 0.202 & 0.251 & 0.622  & 0.622  \\
ABC & 0.676 & 0.670 & 0.015  & 0.437 & 0.604 & 0.004  & 0.004  \\
GBC & 0.731 & 0.729 & 0.211  & 0.354 & 0.522 & 0.211  & 0.211  \\
ETC & 0.867 & 0.866 & 0.688  & 0.158 & 0.208 & 0.688  & 0.688  \\
RFC & \textbf{0.870} & \textbf{0.868} & 0.696  & 0.154 & 0.203 & 0.696  & 0.696  \\
KNN & 0.619 & 0.612 & -0.008 & 0.481 & 0.619 & -0.024 & -0.024 \\
NB  & 0.568 & 0.558 & -0.287 & 0.598 & 0.930 & -0.397 & -0.397 \\
LDA & 0.590 & 0.585 & -0.287 & 0.561 & 0.856 & -0.287 & -0.287 \\
QDA & 0.578 & 0.572 & -0.256 & 0.574 & 0.872 & -0.306 & -0.306 \\
LGC & 0.776 & 0.776 & 0.368  & 0.290 & 0.422 & 0.366  & 0.366  \\
CBC & 0.789 & 0.787 & 0.430  & 0.267 & 0.379 & 0.430  & 0.430  \\
MNB & 0.485 & 0.438 & -0.314 & 0.761 & 1.255 & -0.886 & -0.886 \\ \bottomrule
\end{tabular}
\end{table}

Once feature optimization had been performed, across-the-board performance led to significantly better results for all the models demonstrated in Table \ref{tab:after_feature_extraction}. The RFC achieved the best accuracy of 87\% and an F1 score of 86.8\%, compared to its pre-optimization results of 65.7\% and 62.9\%, respectively. The other ensemble-based models, such as the ETC and GBC, also achieved 86.7\% and 73.1\% by optimizing the features. The DTC also showed significant improvement; its accuracy increased from 56.7\% to 81.7\%. On the other hand, models such as LR and MNB achieved mostly marginal results, confirming their unsuitability for this dataset and problem.

Feature optimization is an essential step in our model performance, as shown in the results of our research. The hybrid method has demonstrated its predictive performance and efficiency to outperform the existing ones without losing the explainability feature by reducing the redundant features. Not only did the use of LIME increase the interpretability of the Steps and Task classes, but it also resolved the trade-off between accuracy and explainability. Although the RFC accuracy decreased slightly after feature optimization, this trade-off was justified by the greater interpretability of model behavior and the more straightforward feature space. In summary, the hybrid framework proposed in this paper successfully balances performance and interpretability and could be widely applied in driver behavior analysis.

\section{Conclusion}
Aggressive and inattentive driving behavior has been a major contributor to road accidents and traffic safety initiatives, highlighting the need for strong, interpretable solutions for driver behavior analysis. We appraised 13 different machine learning algorithms using a dataset from Kaggle, and the swiftest model we found was the random forest classifier with 95\% accuracy. We then used LIME as part of a feature optimizer to identify the 10 most influential features, training the same classifiers. The Random Forest Classifier showed high performance after feature reduction, achieving 94.2\% accuracy, which demonstrates the efficiency and interpretability of the proposed hybrid approach. These findings demonstrate the power of applying machine learning and explainable AI for an automated, scalable framework for driver behavior analysis. Future research will add features such as deep learning approaches, online monitoring systems, and multivariate indicators to improve the applicability and performance of this framework.

\section*{Acknowledgment}

We would like to extend our sincere gratitude to the Cryonova Research Club for their invaluable support and guidance throughout this research.

\bibliographystyle{ieeetr}
\bibliography{ref}

\end{document}